\documentclass[letterpaper]{article} % DO NOT CHANGE THIS
\usepackage[preprint]{aaai2027}  % arXiv preprint: suppress the review-only notice
\usepackage[hyphens]{url}  % DO NOT CHANGE THIS
\usepackage{graphicx} % DO NOT CHANGE THIS
\usepackage{natbib}  % DO NOT CHANGE THIS AND DO NOT ADD ANY OPTIONS TO IT
\usepackage{caption} % DO NOT CHANGE THIS AND DO NOT ADD ANY OPTIONS TO IT
\usepackage{algorithm}
\usepackage{algorithmic}

\usepackage{newfloat}
\usepackage{listings}
\DeclareCaptionStyle{ruled}{labelfont=normalfont,labelsep=colon,strut=off} % DO NOT CHANGE THIS
\floatstyle{ruled}
\newfloat{listing}{tb}{lst}{}
\floatname{listing}{Listing}

\usepackage{booktabs}

\usepackage{amsmath}
\usepackage{amssymb}
\usepackage{amsfonts}
\usepackage{multirow}
\usepackage{subcaption}
\usepackage{makecell}
\usepackage{tikz}

\title{$x$-Prediction Flow: Efficient Continuous Decoding \\
for Masked Diffusion Language Models}
\author{
    Weitian Wang\textsuperscript{\rm 1,\rm 2},
    Lianlei Shan\textsuperscript{\rm 3},
    Shubham Rai\textsuperscript{\rm 1},
    Cecilia De La Parra\textsuperscript{\rm 1},
    Akash Kumar\textsuperscript{\rm 2}
}
\affiliations{
    \textsuperscript{\rm 1}Robert Bosch GmbH, Germany\\
    \textsuperscript{\rm 2}Ruhr University Bochum, Germany\\
    \textsuperscript{\rm 3}University of the Chinese Academy of Sciences, China\\
    weitian.wang@bosch.com, akash.kumar@ruhr-uni-bochum.de
}

\begin{document}

\maketitle

\begin{abstract}
Masked diffusion language models (MDLMs) generate text by iteratively unmasking tokens, but their standard decoder reduces each step to a binary action: a position is either committed to a single token or left fully masked, discarding rich predictive information rather than carrying it forward, and forcing premature, irrevocable commitments that lead to poor performance under a limited decoding budget.
In this paper, we reinterpret mask prediction as a clean-state prediction ($x$-prediction) and show that it can be used to induce a continuous flow in the input embedding space.
Building on this view, we propose a continuous decoding framework for MDLMs where tokens can accumulate partial progress at each diffusion step and remain revisable.
To match the uneven contextual constraints across positions in language, we replace the globally synchronous schedule in image diffusion with a confidence-based asynchronous update in which the diffusion progress is token-wise accumulated. Additionally, we introduce a lightweight policy network and formulate its training as a reinforcement learning problem. Applied to pretrained LLaDA, our decoder retains 83--97\% of full-budget accuracy using under 15\% of the diffusion steps, largely outperforming discrete mask-prediction decoding at matched budgets.
\end{abstract}

\section{Introduction}

%Large language models built on the autoregressive (AR) paradigm~\cite{achiam2023gpt,grattafiori2024llama} have driven recent progress in natural language processing. However, their strictly sequential factorization imposes a fundamental constraint at inference time: tokens must be produced one after another, conditional only on the previously generated prefix. This sequential dependency imposes a high decoding latency. This is further exacerbated in modern reasoning models, where chain-of-thought traces routinely span thousands of tokens~\cite{guo2025deepseek}.

Diffusion language models (DLMs) have recently emerged as a promising alternative~\cite{nie2025large,bie2025llada2} to the autoregressive (AR) paradigm. Instead of committing tokens left-to-right, DLMs cast generation as an iterative denoising process: a fully corrupted response is refined in parallel over a small number of steps, with every position able to attend bidirectionally to the rest of the sequence. This parallelism enables faster generation, supports controllable and non-causal generation patterns, and provides a built-in mechanism for refining earlier predictions in light of later context.

\begin{figure}[t]
\centering
\includegraphics[width=0.45\textwidth]{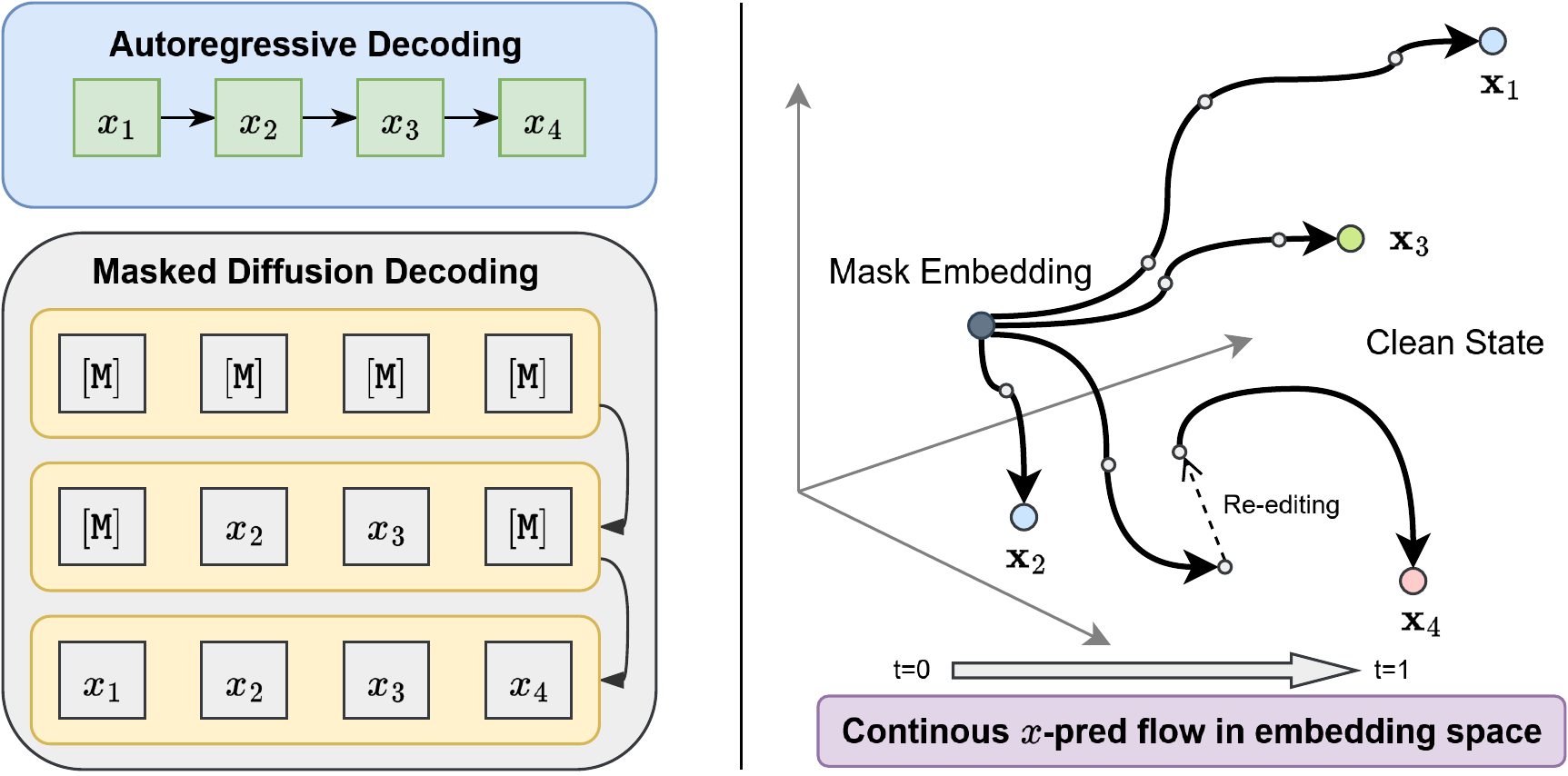}
\caption{By reinterpreting mask prediction as clean-state prediction in embedding space and defining a velocity field that moves the current state from mask embedding toward the predicted clean state, our $x$-prediction flow decoding evolves \emph{all} tokens \emph{continuously} in embedding space.}
\label{fig:decodings}
\vspace{-1em}
\end{figure}

A natural design is to follow image diffusion~\cite{ho2020denoising,lipman2022flow,li2025back} and denoise directly in a continuous embedding space. Language tokens are discrete and highly context-dependent, with the meaning of a token determined by the identities of its neighbours rather than by its position in any continuous coordinate. The dominant DLM family~\cite{ou2024your,nie2025large,bie2025llada2} instead adopts \emph{masked} diffusion, building on the long-standing success of mask prediction popularized by BERT~\cite{devlin2019bert}, in which a fraction of tokens are corrupted with a special $[\texttt{M}]$ symbol and the model is asked to recover these masked tokens from the surrounding context. This formulation has enabled MDLMs to scale to billions of parameters and to match AR baselines on standard benchmarks, while also showing advantages on tasks that benefit from bidirectional structure, such as code generation and reverse reasoning~\cite{nie2025large}.

Despite this scalability, the standard decoding procedure of MDLMs makes inefficient use of its compute budget. At each step, the model outputs a categorical distribution over the vocabulary at every masked position, but the sampler reduces this distribution to a binary action: either the position is unmasked to a single committed token, or it is left as $[\texttt{M}]$ and re-predicted from scratch at the next step. From the perspective of any individual position, the per-step state is therefore all-or-nothing: the model's belief is consumed to make a hard decision and then discarded, rather than carried forward. Two consequences follow. First, the rich predictive information contained in the full output distribution, including the runner-up candidates and their relative likelihoods, is lost as soon as the step ends, and a position left masked begins the next step with no memory of it. Second, once a token is unmasked, the commitment is final, even when later updates to neighbouring positions would have favoured a different choice. The model is forced to choose between premature certainty and complete uncertainty, which limits how effectively a fixed number of decoding steps can be used. This limitation motivates the design of a decoding scheme in which each position maintains an intermediate estimate that is carried across steps and refined \emph{continuously}, so that confidence accumulates gradually with context and committed predictions remain revisable until generation converges.

In this work, we realize this continuous decoding scheme by reinterpreting mask prediction as clean-state prediction. We treat the model's masked-position output as an estimate of the clean state in embedding space, and use it to define a velocity field that moves the current state from mask embedding toward the predicted clean state. We refer to the resulting continuous dynamics as an $x$-prediction flow. Unlike image diffusion, which starts from stochastic Gaussian noise, this flow is initialized from the deterministic mask embedding, a state on which MDLMs are explicitly trained. Building on this flow, we further adapt the schedule to the asymmetric, context-dependent nature of language generation, where some tokens must commit early to inform others. Our main contributions are:
\begin{itemize}
    \item \textbf{A continuous decoding paradigm for pretrained MDLMs.} We show that mask-prediction decoding can be reformulated as an $x$-prediction flow in the input embedding space, where each trajectory starts from the mask embedding and is iteratively moved toward the predicted clean state. This turns binary unmasking into continuous, revisable updates and runs on off-the-shelf MDLMs with only a few hundred alignment-training steps.

    \item \textbf{An asynchronous token-wise diffusion schedule with adaptive compute.} We replace the globally synchronous schedule in image diffusion with a confidence-based asynchronous update scheme in which each token carries its own decoding progress. This allows high-confidence positions to progress faster, providing a clearer context for other tokens. The decoding progress also measures how much of the mask embedding a token still retains, which directly yields a stopping criterion without any additional module: decoding halts once every position has converged, so the number of steps spent adapts to the difficulty of each sample.

    \item \textbf{A learned token-wise step-size policy.} We parameterize each token's update as a log-scale fraction of its remaining diffusion distance, conditioned on the token's confidence statistics and current decoding progress, and train this policy with GRPO~\cite{shao2024deepseekmath} using task-level rewards plus a completion regularization term.
\end{itemize}

\section{Masked Diffusion Language Model}
\label{sec:mdlm}
This section introduces masked diffusion language models (MDLMs)---a family of non-autoregressive generative models for text that recover clean tokens from partially masked sequences---and establishes the notation used throughout the rest of the paper. We describe MDLMs in their general form, of which LLaDA~\cite{nie2025large} is a representative instantiation at scale.

Let $\mathcal{V}$ be a discrete vocabulary augmented with a special mask token $[\texttt{M}]$.
A language sequence of length $N$ is $\mathbf{x}_0 = (x_0^1, \ldots, x_0^N) \in \mathcal{V}^N$.
Each token $v \in \mathcal{V}$ is associated with a learned embedding via an embedding matrix $\mathbf{W}_e \in \mathbb{R}^{|\mathcal{V}| \times E}$, and we denote the mask embedding as $\mathbf{m} \triangleq \mathbf{W}_e[[\texttt{M}]] \in \mathbb{R}^E$.
We assume access to a corruption mechanism that, given a mask ratio $\sigma \in [0,1]$, produces a partially masked sequence $\tilde{\mathbf{x}}$ in which a subset $\mathcal{M} \subseteq \{1,\ldots,N\}$ of positions (with expected size $\sigma N$) are replaced by $[\texttt{M}]$, while the remaining positions $\mathcal{U} = \{1,\ldots,N\} \setminus \mathcal{M}$ retain their original tokens.

\subsubsection{Training objective}
An MDLM is trained to recover the clean tokens at masked positions given the surrounding context. Concretely, the model defines a conditional distribution $p_\theta(x_0^i \mid \tilde{\mathbf{x}})$ for each position $i$, and is optimized by minimizing a cross-entropy loss restricted to the masked positions:
\begin{equation}
    \mathcal{L}_{\text{MDLM}}(\theta) = -\,\mathbb{E}_{\sigma,\, \mathbf{x}_0,\, \tilde{\mathbf{x}}}
    \left[\frac{1}{\sigma}\sum_{i=1}^{N} \mathbf{1}\!\left[i \in \mathcal{M}\right] \log p_\theta\!\left(x_0^i \mid \tilde{\mathbf{x}}\right)\right],
    \label{eq:mdlm_loss}
\end{equation}
where $\sigma \sim \mathcal{U}[0,1]$ and the $\frac{1}{\sigma}$ weighting compensates for the expected fraction of masked tokens. This objective has been shown to be a variational upper bound on the negative log-likelihood of the data distribution~\cite{ou2024your,nie2025large}, providing a principled likelihood-based training criterion.

\subsubsection{Mask predictor}
The conditional $p_\theta(\cdot \mid \tilde{\mathbf{x}})$ is parameterized by a \emph{mask predictor} $f_\theta$, typically realized as a bidirectional Transformer~\cite{vaswani2017attention} so that every masked position can attend to the full surrounding context. Given $\tilde{\mathbf{x}}$, the predictor outputs per-position logits
\begin{equation}
    \mathbf{z}_{\text{pred}}^i = f_\theta(\tilde{\mathbf{x}})^i \in \mathbb{R}^{|\mathcal{V}|},
    \quad
    p_\theta(\cdot \mid \tilde{\mathbf{x}})^i = \operatorname{softmax}(\mathbf{z}_{\text{pred}}^i),
\end{equation}
and the predicted token at position $i$ is $x_{\text{pred}}^i = \arg\max_{v}\, p_\theta(v \mid \tilde{\mathbf{x}})^i$. Crucially, all masked positions are predicted in parallel within a single forward pass, in sharp contrast to the sequential left-to-right factorization of autoregressive language models.

\subsubsection{Standard discrete sampling}
At inference time, the input to the model is the concatenation of a clean prompt $\mathbf{p}$ and a fully masked response of a chosen length, and decoding proceeds over a fixed number of steps until every response position is filled. At each step, $f_\theta$ predicts all currently masked response tokens simultaneously conditioned on $\mathbf{p}$ and the partially decoded response; a subset of these predictions is then committed (unmasked), while the rest are re-masked for the next step. The prompt $\mathbf{p}$ is never masked. Existing MDLMs typically commit a fixed fraction of tokens per step and rely on heuristics such as low-confidence remasking---which retains the most confident predictions and re-masks the least confident ones~\cite{nie2025large}---to decide which tokens to commit.
However, this discrete, synchronous update ignores the continuous structure of the token embedding space and forces premature hard decisions that cannot be revised in later steps---motivating the continuous decoding framework introduced next.

\section{Continuous-State Diffusion for MDLMs}
The goal of this paper is to bring continuous-state diffusion to MDLMs. Unlike the discrete update in standard MDLM decoding, 
% where each masked position is either committed to a single token or left fully masked at every step, 
a continuous-state decoder can carry a soft intermediate estimate of every token across steps and refine it gradually. Preserving this intermediate state instead of collapsing it into a hard decision uses each diffusion step more efficiently and yields higher-quality generations under a limited diffusion budget. To this end, we propose a continuous decoding framework based on an $x$-prediction flow in token embedding space, anchored at the mask embedding rather than at Gaussian noise, therefore letting a pretrained MDLM be operated as a continuous-state diffusion model with only lightweight alignment training.

\subsection{Challenges of Transferring Continuous-State Diffusion to Language}
\label{sec:challenges}
% In this section, we first identify two fundamental challenges of transferring continuous-state diffusion from images to language, which will be addressed by our method in the subsequent subsections. 
Continuous-state diffusion has achieved remarkable success in the image domain~\cite{ho2020denoising,lipman2022flow,li2025back}, where a clean state is recovered by gradually denoising a corrupted continuous state through a sequence of small refinement steps. Naively transplanting this recipe to language, however, runs into two fundamental obstacles that any continuous decoder for text must confront.

\subsubsection{Gaussian noise is not a well-defined state for pre-trained MDLMs}
\label{sec:gaussian_noise}
In continuous-state image diffusion~\cite{ho2020denoising,lipman2022flow,li2025back}, the corruption process involves adding some kind of Gaussian noise to pixel values, which represents a smooth change to the original value and preserves spatial structures. On the other hand, adding Gaussian noise to a token significantly alters its semantic meaning, making the difference between original and corrupted state absolute, which leads reverse diffusion trajectories in MDLMs to map their predictions onto completely different, arbitrary tokens \cite{gao2024empowering}.

\subsubsection{Language generation is highly context-dependent}
Continuous-state image diffusion~\cite{ho2020denoising,lipman2022flow,li2025back} adopts a synchronous denoising schedule because pixels % although locally correlated, 
can be refined largely in parallel: most of the information needed to disambiguate one pixel is shared symmetrically with its neighbors and is gradually revealed as noise is removed. Such synchronous schedule cannot be applied to language, which exhibits a much stronger and more asymmetric form of contextual dependence. The identity of a token can be determined almost entirely by a few specific tokens elsewhere in the sequence, and those informative tokens may themselves be uncertain. A useful continuous decoder must therefore allow asymmetric refinement enabling some positions to commit early and serve as anchors for the rest, while leaving other positions open to refinement until enough surrounding context has stabilized. %A globally synchronous schedule with a single shared step size, as in standard image diffusion, cannot express this asymmetric refinement order and forces every position to be resolved at the same rate regardless of how informative its current context is.<--this is repeating what is stated in the first lines of this subsection

\subsection{Observation: Clean Token Prediction Consistency}
\label{sec:selfconsist}
We encounter a third obstacle concerning the model's behavior once a state has nearly converged. A continuous decoder maintains a soft estimate at every position and re-queries the model at each step, including at positions whose state is already close to a clean token embedding. Standard MDLM sampling never issues such a query, since a committed position is never read again. Under a continuous decoder the prediction at a near-clean position is consumed at every subsequent step and therefore influences subsequent prediction dynamics.

To characterize this behavior, we probe LLaDA on the $4{,}763$ clean response positions of $64$ MBPP problems. We present the fully clean sequence as input and compare the predicted embedding---the predicted vocabulary distribution read out against the embedding table---with the embedding of the input token.
In terms of arg max token id correspondence, the model reproduces the input token $96.5\%$ of the time. In terms of distance in the embedding space, however, the readout sits a relative $24\%$ away in $L_2$ norm, far more than the $3.5\%$ token error rate alone would account for: probability mass on near-synonyms displaces the readout even where the arg max is correct.

These observations show that pretrained LLaDA implicitly performs highly consistent clean-signal prediction at clean response positions, making it a sound initialization for our designs in the following sections.

\subsection{$x$-Prediction Flow Anchored at Mask Embedding}
\label{sec:x_pred}
To address the first challenge posed in Sec.~\ref{sec:gaussian_noise}, we anchor the corrupted state at $\mathbf{m}$ rather than at Gaussian noise, as MDLMs are pre-trained to denoise sequences by transitioning specifically from mask embeddings to clean tokens. We formulate our corrupted state $\mathbf{X}_{\text{in}} \in \mathbb{R}^{N\times E}$ as:
\begin{equation}
\mathbf{X}_{\text{in}} = \mathbf{t}\cdot \mathbf{X}_{0} + (1-\mathbf{t})\cdot \mathbf{m}
\label{eq:x_in}
\end{equation}
where N is the number of response tokens, and E is the embedding dimension. $\mathbf{X}_0$ denotes some clean tokens, and $\mathbf{m}$ is the mask embedding that represents a well-defined unknown state for the pretrained MDLMs, and $\mathbf{t}$ is the diffusion time, which represents the corruption level in diffusion and decoding progress in our case.

At the beginning of the diffusion process, where $\mathbf{t}=0$, the model starts with pure mask embeddings, which align with the training setup of MDLMs. During the diffusion process, the model refines the response and increases the decoding progress $\mathbf{t}$ at each step. When $\mathbf{t}$ reaches 1, the tokens are fixed to the final believed clean tokens $\mathbf{X}_0$.

At each diffusion step, the MDLM $f_{\theta}$ produces per-position logits over the vocabulary from the corrupted input. We take the argmax token at every position and map it back into the input embedding space by looking up its row in the embedding matrix $\mathbf{W}_e$ to obtain the predicted state $\mathbf{X}_{\text{pred}}$:
\begin{equation}
\mathbf{X}_{\text{pred}}^i = \mathbf{W}_e\bigl[\,\operatorname*{argmax}_{v \in \mathcal{V}}\, f_{\theta}(\mathbf{X}_{\text{in}})^i_v\,\bigr],
\qquad i = 1,\ldots,N,
\label{eq:hard_readout}
\end{equation}
where the arg max is taken over the vocabulary dimension independently at each position $i$. Based on the findings in Sec.~\ref{sec:selfconsist}, we interpret this prediction as the current estimate of the clean signal in the input space, the $x_0$-prediction parameterization of diffusion models~\cite{ho2020denoising,karras2022elucidating}. We follow JiT~\cite{li2025back} and define the velocity field as
\begin{equation}
    \mathbf{V}(\mathbf{X}_{\text{in}},\mathbf{t}) = \frac{\mathbf{X}_{\text{pred}} - \mathbf{X}_{\text{in}}}{\max(1-\mathbf{t},\; 0.05)},
    \label{eq:velocity}
\end{equation}
where the denominator is clamped from below at $0.05$, elementwise, so that the velocity stays bounded as $\mathbf{t} \rightarrow 1$. We then update the current state $\mathbf{X}_{\text{in}}$ with
\begin{equation}
    \mathbf{X}_{\text{in}} = \mathbf{X}_{\text{in}} + \mathbf{V}\cdot \Delta \mathbf{t}.
    \label{eq:euler}
\end{equation}
This formulation allows each token to be continuously refined in the input space of the MDLM.

\subsection{Confidence-based Asynchronous Decoding}
\label{sec:async_decode}
As discussed in Sec.~\ref{sec:challenges}, language generation is highly contextually dependent, and thus we need to fix some tokens first to infer other tokens. To meet this nature of language generation, we introduce a confidence-based asynchronous update mechanism where the decoding progress $t$ is a vector $\mathbf{t} \in \mathbb{R}^N$.

The step size is defined as:
\begin{equation}
\Delta \mathbf{t} = \mathbf{a} \cdot (\mathbf{1}-\mathbf{t})
\label{eq:step}
\end{equation}
where $\mathbf{a} \in [a_{\min}, a_{\max}]^N$ is a per-token step fraction produced by a confidence-based step-size policy that takes the model's predictive distribution and the current decoding progress as input. The policy parameterization and training are detailed in Sec.~\ref{sec:rl}. After each step, the decoding progress $\mathbf{t}$ is updated with:
\begin{equation}
 \mathbf{t} = \mathbf{t} + \Delta \mathbf{t}
\end{equation}
This mechanism allows high-confidence tokens to progress faster, providing a richer context for the refinement of low-confidence tokens.

Combining Eq.~\ref{eq:velocity}, Eq.~\ref{eq:euler} and Eq.~\ref{eq:step}, and setting the clamp aside, the continuous update takes the compact form
\begin{equation}
    \mathbf{X}_{\text{in}} \leftarrow (\mathbf{1}-\mathbf{a}) \odot \mathbf{X}_{\text{in}} + \mathbf{a} \odot \mathbf{X}_{\text{pred}},
    \label{eq:collapse}
\end{equation}
where $\odot$ broadcasts the per-token gain over the embedding dimension: the factor $1-\mathbf{t}$ cancels and the progress state no longer appears explicitly. This does not make the schedule trivial, because $\mathbf{a}$ is itself a function of $\mathbf{t}$---it is produced by a policy conditioned on the current decoding progress $\mathbf{t}$ and the model's confidence (Sec.~\ref{sec:rl})---so the gain applied at a position depends on how far that position has already advanced. Setting $a_i = \Delta t / (1 - t_i)$ at every position, for a single globally shared scalar $\Delta t$, makes Eq.~\ref{eq:step} return the same increment $\Delta t$ everywhere and recovers the fixed-step-size Euler update of JiT~\cite{li2025back}. Our update is therefore a strict generalization of it, in which the single shared schedule is replaced by a per-token, confidence-conditioned one.

\subsubsection{Early stopping}
\label{sec:earlystop}
As shown in Eq.~\ref{eq:x_in}, $\mathbf{t}$ can serve as an explicit signal of how far each token still has to travel: $1 - t_i$ is the weight the mask embedding $\mathbf{m}$ retains in the current state of token $i$.
Early stopping therefore needs no separate criterion: decoding stops when \emph{all} tokens have crossed a threshold $\tau$,
\begin{equation}
\min_i t_i \geq \tau,
\label{eq:earlystop}
\end{equation}
with $\tau = 0.9$ in all our experiments. Once Eq.~\ref{eq:earlystop} holds, every token has retained at most $1 - \tau$ of the mask embedding, and further iterations would only refine a state whose nearest-token readout has already settled.

Stopping this way makes the decoding budget adaptive without any explicit scheduling. The step-size policy assigns a small $a_i$ where the model's predictive distribution is uncertain, so by Eq.~\ref{eq:step} that token's decoding progress advances slowly. Hence, a hard sample that contains many uncertain positions requires more decoding steps than an easy one, where every token reaches a high decoding progress fast. This further reduces the number of steps our method takes on average, as shown in Table~\ref{tab:code_gen}.

\subsubsection{Stochasticity of the decoder}
\label{sec:stochasticity}
The decoding procedure described above is not inherently deterministic. Two independent sources of randomness enter every step. The first is the clean-state estimate: $\mathbf{X}_{\text{pred}}$ is a readout of the MDLM's own predictive distribution, so the direction of the velocity field in Eq.~\ref{eq:velocity} is itself random. The second is the step size: the action $\mathbf{a}$ in Eq.~\ref{eq:step} is sampled from the step-size policy, which is a Beta distribution over the latent variable (Sec.~\ref{sec:rl}), so the distance travelled along that direction is random as well. Because the two interact through the confidence statistics that condition the policy, the decoder as a whole defines a distribution over output sequences rather than a single trajectory.

During evaluations, we use greedy decoding for MDLMs, and for the step-size policy we replace the sampled action by its expectation under the transformed Beta distribution, as described in Sec.~\ref{sec:rl}.

\subsection{Confidence-Driven Discrete Adjustments}
Standard MDLM decoding makes two confidence-driven decisions at every step: it commits the most confident predictions and re-masks the rest. On top of our continuous flow, we retain two analogous discrete adjustments that inherit the same confidence signal: one rolls back the diffusion state when a previously high-confidence prediction is no longer trusted, and the other commits the most reliable prediction early to anchor the rest. We define the confidence of token $i$ as
\begin{equation}
\label{eq:conf}
c_i = \operatorname{max}(\operatorname{softmax}(\mathbf{z}_i)),
\end{equation}
where $\mathbf{z}_i$ is the output logits of token $i$.

\subsubsection{Re-editing}
The step size $\Delta \mathbf{t}$ in Sec.~\ref{sec:async_decode} is positive, so decoding progress is monotonically increasing. In practice, the model is sometimes no longer confident in a token whose progress has already advanced. To allow such tokens to be re-estimated, when $c_i - t_i < -0.1$ at a given step we re-edit the state of token $i$ as a mixture of the current prediction $\mathbf{x}_{\text{pred}}$ and the mask embedding $\mathbf{m}$:
\begin{equation}
\mathbf{x}_i = c_i \cdot \mathbf{x}_{\text{pred}} + (1-c_i) \cdot \mathbf{m},
\end{equation}
and reset its decoding progress to $t_i = c_i$. Intuitively, when the model's confidence drops well below the position's commitment level, we let the diffusion state jump back toward the current estimation and abandon the previously trusted information.

\subsubsection{Hard commitment}
At each diffusion step, among all tokens with decoding progress $t_i < 0.99$, we fix the token with the highest confidence and set its decoding progress to $1$. This anchors the most reliable prediction so that it provides a stable context for the remaining tokens, mirroring the commit-by-confidence heuristic of standard MDLM decoding.

\section{Training}
In this section, we describe how we adapt a pretrained MDLM for $x$-prediction flow and how we train the step-size policy.

\subsection{$x$-Prediction Alignment}
\label{sec:alignment}

The consistency observed in Sec.~\ref{sec:selfconsist} is a behaviour that MDLM training never constrains. As discussed in Sec.~\ref{sec:mdlm}, the MDLM objective minimizes a cross-entropy loss at \emph{masked positions} only: the indicator $\mathbf{1}[i \in \mathcal{M}]$ in Eq.~\ref{eq:mdlm_loss} leaves the output at unmasked positions entirely free. In this section we make clean token prediction consistency an explicit training target, so as to reduce the embedding-space deviation observed in Sec.~\ref{sec:selfconsist}.

When we reinterpret mask prediction as $x$-prediction flow as stated in Sec.~\ref{sec:x_pred}, we need the model to:
\begin{enumerate}
    \item Predict clean token $\mathbf{x}_0^i$ from $\mathbf{m}$ on all masked positions $i \in \mathcal{M}$.
    \item Predict the identical clean token $\mathbf{x}_0^i$ from $\mathbf{x}_0^i$ on all unmasked positions $i \in \mathcal{U}$ because the estimation on clean diffusion states should remain the same in a stable dynamic system.
\end{enumerate}
% Pretraining already supplies requirement 1; requirement 2 is what alignment must add. We therefore optimize both in a single embedding-space objective, which enforces requirement 2 at unmasked positions while holding requirement 1 in place at masked ones.

\noindent For any masked input sequence $\mathbf{X}_{\text{in}} =
\begin{cases}
\mathbf{m}, & \text{if masked} \\
\mathbf{X}_0, & \text{otherwise}
\end{cases}$ and MDLM $f_\theta$,
we construct the following loss:
\begin{equation}
    \mathbf{X}_{\text{pred}} = \operatorname{softmax}_{\mathcal{V}}(\mathbf{Z}_{\text{pred}})\, \mathbf{W}_{\text{in}}, \qquad
    \mathbf{Z}_{\text{pred}} = f_{\theta}(\mathbf{X}_{\text{in}}),
    \label{eq:soft_readout}
\end{equation}
where $\operatorname{softmax}_{\mathcal{V}}$ is applied over the vocabulary dimension independently at each position,
\begin{equation}
    \mathcal{L}_{x\text{-pred}} = \left\| \mathbf{X}_{\text{pred}} - \mathbf{X}_0 \right\|_2^2
\end{equation}

The $x$-prediction loss $\mathcal{L}_{x\text{-pred}}$ constrains the model to predict clean states from corrupted sequences in the embedding space without hurting its original mask prediction ability, which we verify in Sec.~\ref{sec:align_exp} by tracking the masked-position cross-entropy against the pretrained reference throughout alignment (Fig.~\ref{fig:ce_losses}). Since we use mean squared error loss instead of cross-entropy loss to explicitly align the model's output in the embedding space, $\mathcal{L}_{x\text{-pred}}$ is not suitable for guiding the model's generation on unseen inputs. We use self-generated answers as clean data $\mathbf{X}_0$ in experiments to make sure that the loss originates from misalignment but not wrong prediction.

\subsection{Step-Size Policy Training}
\label{sec:rl}
As discussed in Sec.~\ref{sec:async_decode}, decoding requires a step-size policy that produces an adaptive step size for each token from its confidence statistics. We train this policy with reinforcement learning because we want it to optimize for end-task performance rather than per-token matching with a reference, and the former is non-differentiable through the multi-step decoder. Among RL algorithms, we adopt GRPO~\cite{shao2024deepseekmath} to avoid learning a separate value baseline, which would be unreliable here since per-step confidence statistics are only weakly predictive of the final sequence reward.

At each decoding step, for each token, we construct a compact policy state from the MDLM predictive distribution and the current diffusion progress:
\begin{equation}
\mathbf{s} =
\left[p_{(1)}, p_{(2)}, p_{(3)}, p_{(4)}, H_k, p_m, t, \rho\right],
\end{equation}
where \(p_{(j)}\) denotes the \(j\)-th largest probability from
\(p = \operatorname{softmax}(\mathbf{z}_{\text{pred}})\), and
\(p_m = p_{(1)} - p_{(2)}\) is the confidence margin. The scalar \(H_k\)
is the normalized entropy over the top-\(k\) probabilities:
\begin{equation}
\begin{aligned}
\tilde{p}_i &= \frac{p_i}{\sum_{j \in S_k} p_j}, \quad i \in S_k, \\
H_k &= - \frac{1}{\log k} \sum_{i \in S_k} \tilde{p}_i \log \tilde{p}_i.
\end{aligned}
\end{equation}
where \(S_k\) is the set of top-\(k\) indices; in our implementation
\(k=4\). Finally, \(t \in [0,1]\) denotes the token-wise decoding progress,
and \(\rho = \frac{\ell}{T-1}\) denotes the normalized global diffusion step
at step \(\ell\) out of \(T\) total steps.

\subsubsection{Log-scale Beta policy}
The step policy parameterizes a Beta distribution over a latent variable \(y \in [0,1]\), which is later mapped to the action \(a\). Concretely, a two-layer MLP with SiLU activations encodes \(\mathbf{s}\) into a hidden representation \(\mathbf{h}_\theta(\mathbf{s}) \in \mathbb{R}^d\), and two linear heads predict the Beta mean and concentration from it:
\begin{equation}
\begin{aligned}
\mu_\theta(\mathbf{s}) &= 0.05 + 0.9 \cdot \sigma\!\left(\mathbf{W}_\mu \mathbf{h}_\theta(\mathbf{s}) + b_\mu\right), \\
\kappa_\theta(\mathbf{s}) &= \kappa_{\min} + (\kappa_{\max} - \kappa_{\min}) \cdot \sigma\!\left(\mathbf{W}_\kappa \mathbf{h}_\theta(\mathbf{s}) + b_\kappa\right),
\end{aligned}
\end{equation}
where \(\mathbf{W}_\mu, \mathbf{W}_\kappa \in \mathbb{R}^{1 \times d}\) and \(b_\mu, b_\kappa \in \mathbb{R}\) are the parameters of the two heads, and \(\sigma\) is the sigmoid. The affine rescalings bound \(\mu_\theta(\mathbf{s}) \in [0.05, 0.95]\) and \(\kappa_\theta(\mathbf{s}) \in [\kappa_{\min}, \kappa_{\max}]\), keeping the Beta distribution away from degenerate endpoints and extreme concentrations while still allowing the policy to represent both conservative and aggressive updates. The latent variable is then sampled as
\begin{equation}
\begin{aligned}
y &\sim \operatorname{Beta}(\alpha_\theta, \beta_\theta), \\
\alpha_\theta &= \mu_\theta(\mathbf{s}) \kappa_\theta(\mathbf{s}), \quad
\beta_\theta = \left(1-\mu_\theta(\mathbf{s})\right)\kappa_\theta(\mathbf{s}).
\end{aligned}
\end{equation}
Instead of using \(y\) directly
as the step fraction, we map it to an action \(a \in [a_{\min}, a_{\max}]\)
on a logarithmic scale:
\begin{equation}
a = 2^{u},
\qquad
u =
\log_2 a_{\min}
+ y\left(\log_2 a_{\max} - \log_2 a_{\min}\right).
\end{equation}
We use \(a_{\min}=1/256\) and \(a_{\max}=1\). The resulting token-wise diffusion step for token $i$ is $\Delta t_i = a_i \cdot (1-t_i)$ as in Eq.~\ref{eq:step}.
Thus, the policy controls the fraction of the remaining diffusion distance to consume at this step. The log-scale action space is important because useful updates span orders of magnitude: uncertain tokens often need very small refinements, while high-confidence tokens can safely consume most of their remaining distance. During deterministic decoding, we replace the sampled action by its policy expectation under this transformed Beta distribution.

\subsubsection{Reward design}
For each prompt \(q_b\) (indexed by \(b\) in the batch), we sample a group of \(G\) decoded trajectories with the current policy, indexed by \(g \in \{1,\ldots,G\}\). Each token receives a reward that combines a trajectory-level task term (e.g., pass rate) and a per-token completion term:
\begin{equation}
R_{b,g,i} = R^{\text{task}}_{b,g} + \lambda \cdot t^{\text{final}}_{b,g,i},
\end{equation}
where \(R^{\text{task}}_{b,g}\) is the trajectory-level task reward and is broadcast to every token in trajectory \(g\), while \(t^{\text{final}}_{b,g,i} \in [0,1]\) is the final decoding progress of token \(i\). The completion term constrains the policy to push every token's diffusion progress toward \(1\), preventing it from leaving tokens partially decoded at the end of the budget.

For the within-group advantage construction and the clipped policy update, we follow GRPO~\cite{shao2024deepseekmath}.

\section{Experiments}
We evaluate our $x$-prediction flow decoding on the code generation and math tasks using LLaDA~\cite{nie2025large}. To show that our method scales up to larger models, we further evaluate on LLaDA2.0~\cite{bie2025llada2}.

\subsection{$x$-Prediction Alignment Training}
\label{sec:align_exp}
For $x$-prediction alignment, we first generate self-alignment data from the Tulu3~\cite{lambert2024tulu}-SFT-Personas-Code dataset using LLaDA and LLaDA2.0. We then train the model with the alignment objective described in Sec.~\ref{sec:alignment}: LLaDA is updated with full-parameter training, while LLaDA2.0 is updated with LoRA~\cite{hu2022lora}. We update LLaDA for 100 steps and LLaDA2.0 for 400 steps. In all experiments below, this alignment training is the only update applied to the pretrained MDLM. The model is not trained on any intermediate states.

\begin{figure}[t]
\centering
\begin{subfigure}[t]{0.48\columnwidth}
    \centering
    \includegraphics[width=\linewidth]{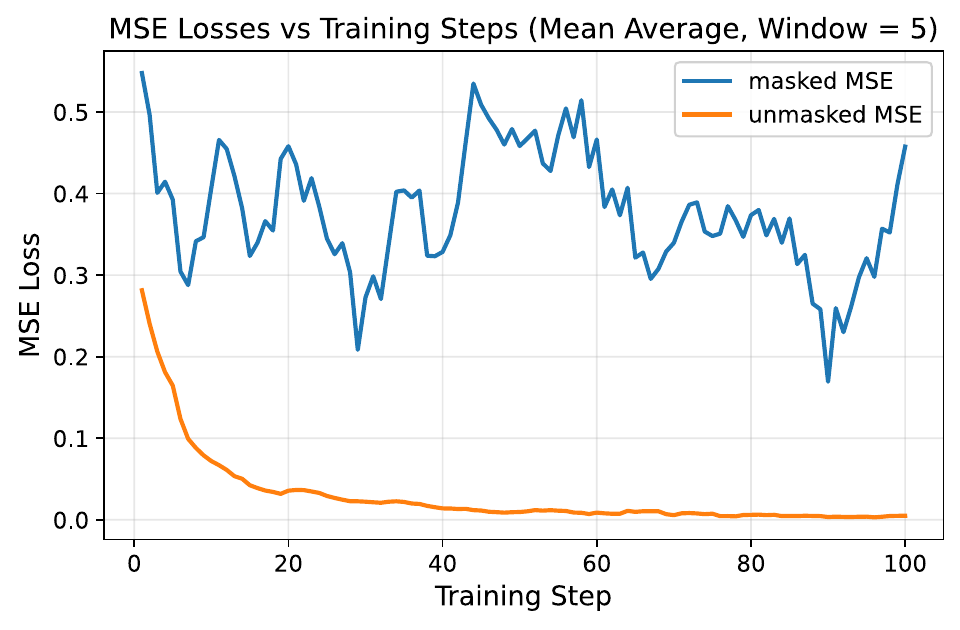}
    \caption{Mean squared error~(MSE) loss.}
    \label{fig:mse_losses}
\end{subfigure}
\hfill
\begin{subfigure}[t]{0.48\columnwidth}
    \centering
    \includegraphics[width=\linewidth]{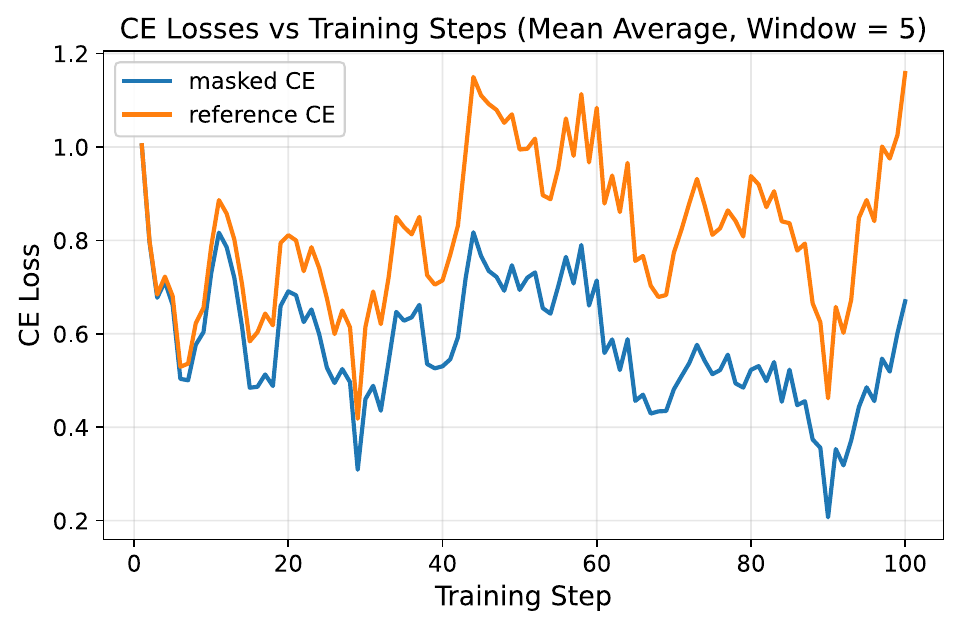}
    \caption{Cross entropy~(CE) loss.}
    \label{fig:ce_losses}
\end{subfigure}
\caption{Training losses during $x$-prediction alignment. The MSE curve tracks masked and unmasked positions, while the CE curve compares the aligned model against the pretrained LLaDA reference at masked positions.}
\label{fig:align_training_losses}
\end{figure}

Fig.~\ref{fig:mse_losses} shows the MSE alignment loss at masked and unmasked positions. After only 100 update steps, the loss on unmasked positions drops close to zero, indicating that the model quickly learns to preserve the clean embedding state when the input token is already clean. Fig.~\ref{fig:ce_losses} further shows that this alignment does not degrade the original masked-token prediction behavior: the CE loss at masked positions remains close to that of the pretrained LLaDA reference.

These results support the central premise of our formulation. Clean-state prediction in embedding space is not a new capability that must be learned from scratch; rather, it is largely compatible with the representations already learned by the MDLM backbone. The lightweight alignment mainly calibrates the model to output a stable clean state at every position, including positions that are not masked during standard MDLM training. Consequently, our reinterpretation of mask prediction as $x$-prediction flow can reuse pretrained MDLMs with minimal backbone modification.

\subsection{GRPO Training}
We train the step-size policy on top of LLaDA-8B-Instruct, using the MBPP~\cite{austin2021program} training split as the prompt source. Each rollout generates a response of length $256$ over $T=256$ diffusion steps.

\subsubsection{Prompt filtering}
Rather than training on the full MBPP training split, we first run the pretrained LLaDA-8B-Instruct on every training problem and keep only the $164$ problems that it can already solve. The step-size policy controls only how the diffusion progress accumulates across tokens and steps, not what the underlying model can solve. For problems that the base model cannot solve under any schedule, every trajectory in a group receives zero task reward; consequently, the within-group advantage collapses and the policy receives no useful learning signal. Restricting training to solvable problems therefore concentrates the reward signal on the regime in which the schedule actually matters.

\subsubsection{Optimization}
We train the policy with a batch size of $8$ prompts and a group size of $G=8$ rollouts per prompt for $6$k steps.

\subsection{Evaluations}
\subsubsection{Setup}
We evaluate on HumanEval~\cite{chen2021codex}, MBPP and GSM8K~\cite{cobbe2021gsm8k}. The generation length is $512$ on HumanEval/GSM8K and $256$ on MBPP. For LLaDA-8B-Instruct, we use a block length equal to the full generation length, i.e., no semi-autoregressive partitioning. LLaDA2.0-mini was pretrained with a blockwise causal mask, so we keep its native semi-autoregressive setup and use a block size of $32$.

\begin{table*}[t]
\centering
\caption{Accuracy~(\%) on HumanEval, MBPP and GSM8K. The budget is a \emph{cap} on diffusion steps (as a fraction of generation length), not the number consumed: our decoder halts once every token reaches $t \geq 0.9$, so \emph{steps used} reports the steps actually executed. Mask prediction has no such exit, so the last block re-runs it at our executed step count.}
\label{tab:code_gen}
\footnotesize
\setlength{\tabcolsep}{4pt}
\begin{tabular}{clccccc}
\toprule
& & \multicolumn{3}{c}{LLaDA-8B-Instruct} & \multicolumn{2}{c}{LLaDA2.0-mini (16B)} \\
\cmidrule(lr){3-5} \cmidrule(lr){6-7}
decoding budget & Method & HumanEval & MBPP & GSM8K & HumanEval & MBPP \\
\midrule
1/1                  & discrete mask prediction & 46.34  & 39.60 & 69.22 & 79.88 & 63.20 \\
\midrule
\multirow{4}{*}{$\leq 1/4$} & discrete mask prediction & 33.54 & 21.20 & 59.67 & 32.32 & 31.40 \\
                     & $x$-prediction flow   & \textbf{45.12} & \textbf{33.00} & \textbf{62.02} & \textbf{59.76} & \textbf{43.80} \\
                     & \quad\emph{avg. steps used} & \emph{73/128} & \emph{35/64} & \emph{69/128} & \emph{128/128} & \emph{64/64} \\
                     & \quad\emph{$\Delta$ vs.\ mask pred.} & \emph{+11.58} & \emph{+11.80} & \emph{+2.35} & \emph{+27.44} & \emph{+12.40} \\
\midrule
\multirow{2}{*}{\makecell[c]{step-matched}} & mask pred.\ @ our steps & 11.59 & 15.20 & 40.26 & -- & -- \\
                     & \quad\emph{$\Delta$ vs.\ $x$-pred.\ flow} & \emph{--33.53} & \emph{--17.80} & \emph{--21.76} & \emph{--} & \emph{--} \\
\bottomrule
\end{tabular}
\end{table*}

\subsubsection{Results}
Table~\ref{tab:code_gen} reports accuracy under a $1/4$ decoding budget, with the full-budget mask-prediction baseline for reference. Our method uses the early stopping mechanism described in Sec.~\ref{sec:async_decode}, so the reported accuracies of $x$-prediction flow are obtained at the \emph{avg. steps used} count, and the step-matched block evaluates mask prediction with the same decoding steps.

$x$-prediction flow improves over mask prediction at the same $1/4$ budget on every dataset. On HumanEval with LLaDA-8B-Instruct, it reaches $45.12$, recovering $97\%$ of the full-budget baseline ($46.34$) with only 14\% of the full budget. On the 16B LLaDA2.0-mini, $x$-prediction flow also brings large improvement over the two datasets.

The margin is smaller on mathematical reasoning: GSM8K improves from $59.67$ to $62.02$ ($+2.35$), recovering $90\%$ of full-budget accuracy against $86\%$ for discrete mask prediction. However, with early stopping, $x$-prediction flow runs only $69$ decoding steps on average. Given the same $69$ steps, discrete mask prediction reaches only $40.26$. Thus, under equal executed compute, the margin is $21.76$ points rather than $2.35$. The compute-matched gains are similarly substantial on code: $11.59$ versus $45.12$ on HumanEval and $15.20$ versus $33.00$ on MBPP.

Both the GSM8K numbers and the LLaDA2.0-mini numbers use the policy trained on MBPP with LLaDA-8B-Instruct (Sec.~\ref{sec:rl}) without retraining, so they are cross-domain and cross-model transfer results, indicating that the policy learns general patterns over per-token confidence statistics rather than domain- or model-specific shortcuts.

\subsection{Comparison with Soft-Masked Decoding}
\label{sec:sm_compare}
In this section, we compare with Soft-Masked Diffusion (SM)~\cite{hersche2025soft}, our closest prior effort that replaces the embedding of every retained mask with a convex combination of the mask embedding and the top-$k$ predicted token embeddings from the previous step, weighted by their confidence scores. SM softens the input representation of retained masks, but its decoding process still follows a discrete position-level schedule: each position is either committed to a token or treated as masked, and the soft mixture is \emph{recomputed} from the current step rather than carried as a persistent continuous state.

SM requires fine-tuning the backbone before its decoder can exploit the soft mixtures. For a fair comparison, we fine-tuned a LLaDA-8B-Instruct model with SM's procedure and then evaluated it both under the SM inference scheme (denoted as soft masking) and under our x-alignment and $x$-prediction flow decoding scheme (denoted as $x$-prediction flow).
The step-size policy is not retrained on the fine-tuned weights. As Table~\ref{tab:sm} shows, SM fine-tuning slightly hurts LLaDA's original performance (cf.\ Table~\ref{tab:code_gen}). We speculate that this is caused by the SM's fine-tuning procedure, which implicitly imposes a bias towards the dataset used, which ideally should be the original dataset used by the model for supervised fine-tuning (SFT). Unfortunately, LLaDA's original SFT dataset is not publicly available.

At the $1/4$ budget, soft masking improves over discrete mask prediction on both datasets. Even on SM's own fine-tuned weights, however, our decoder still surpasses it by $+9.14$ on HumanEval ($37.80$ vs.\ $28.66$) and $+10.60$ on MBPP ($33.80$ vs.\ $23.20$).

\begin{table}[t]
\centering
\caption{Comparison with soft-masked decoding on SM's fine-tuned checkpoint. All rows use the same weights; only the decoding rule differs.}
\label{tab:sm}
\begin{tabular}{clcc}
\toprule
Budget & Decoding Method & HumanEval & MBPP \\
\midrule
1/1 & discrete mask prediction & 44.51 & 36.20 \\
\midrule
\multirow{3}{*}{1/4} & discrete mask prediction & 22.56 & 18.60 \\
                     & soft masking & 28.66 & 23.20 \\
                     & $x$-prediction flow (ours) & \textbf{37.80} & \textbf{33.80} \\
\bottomrule
\end{tabular}
\end{table}

\subsection{Ablations}
\label{sec:ablations}
We ablate the main design components on LLaDA-8B-Instruct using HumanEval under the $1/4$ decoding budget. Table~\ref{tab:ablations} reports the accuracy after removing one component at a time. The result shows that all our design components are important for the final performance, especially $x$-prediction alignment.

\begin{table}[t]
\centering
\caption{Ablation results on HumanEval with LLaDA.}
\label{tab:ablations}
\setlength{\tabcolsep}{4pt}
\begin{tabular}{lcccc}
\toprule
 & Full & \makecell{w/o Hard\\commitment} & \makecell{w/o Re-\\editing} & \makecell{w/o $x$-pred.\\Alignment} \\
\midrule
Accuracy & \textbf{45.12} & 41.46 & 26.83 & 17.07 \\
$\Delta$ & -- & -3.66 & -18.29 & -28.05 \\
\bottomrule
\end{tabular}
\end{table}

\vspace{-1em}
\section{Conclusion}
We introduced $x$-prediction flow, a continuous decoding framework that reinterprets mask prediction in MDLMs as clean-state prediction in embedding space. By anchoring the flow at the mask embedding, maintaining token-wise decoding progress, and learning confidence-conditioned step sizes, our method turns binary unmasking into continuous, revisable refinement while remaining compatible with pretrained MDLMs. Experiments on code generation and math tasks show that this formulation substantially improves low-budget decoding, and ablations confirm the importance of our design components.

\bibliography{aaai2027}
\clearpage
\appendix

\onecolumn
\begin{center}
    {\LARGE\bfseries Appendix}
\end{center}
\vspace{1em}
\section{Related Work}
\label{sec:related_work}

\subsubsection{Few-step decoding for masked diffusion language models}
Although MDLMs predict all masked positions in parallel, high-quality generation can still require many model evaluations because only a subset of predictions is accepted at each step. Self-Distillation Through Time (SDTT)~\cite{deschenaux2025sdtt} compresses a long discrete diffusion trajectory into a much shorter one, exchanging additional distillation for fewer inference steps. dParallel~\cite{chen2026dparallel} instead uses certainty-forcing distillation so that more masked positions become reliable simultaneously and can be committed in parallel. ReMDM~\cite{wang2025remdm} modifies the reverse process to remask previously generated tokens, restoring iterative correction and enabling inference-time scaling while retaining discrete token states. These methods address decoding efficiency through the training target or the discrete transition rule. Our approach modifies a different design choice: it replaces the hard mask-or-committed-token state used by these decoding rules with a persistent embedding-space state, maintains a scalar progress variable per token, and learns the corresponding step-size policy. It could potentially be combined with some acceleration or remasking methods, but we do not evaluate such combinations.

\subsubsection{Soft and revisable decoding}
Among existing masked-diffusion decoders, Soft-Masked Diffusion Language Models (SM)~\cite{hersche2025soft} are the most directly related to our state representation. For every position that remains masked, SM blends the mask embedding with the top-$k$ predicted token embeddings from the preceding step, thereby carrying more information than a hard mask alone. Its position-level schedule nevertheless remains discrete: a position is either committed or retained as a soft mask, and SM recomputes each uncommitted mixture from the latest categorical prediction rather than treating the preceding embedding as the state to be updated. In contrast, our decoder makes the intermediate embedding itself persistent, updates an explicit token-wise progress variable across iterations, and, when confidence falls, resets both the progress and state toward the mask through re-editing. We therefore compare the two decoding rules directly in Sec.~\ref{sec:sm_compare} under the same backbone and budget. EvoToken-DLM~\cite{evotoken2026} likewise replaces hard masks with evolving soft token distributions and supports revision, but learns these dynamics with continuous trajectory supervision. Our method instead performs embedding-space refinement on a pretrained MDLM without training on intermediate trajectory states; we train only endpoint alignment and a token-wise step-size policy.

\subsubsection{Continuous diffusion language models}
Continuous-state generation for text predates our work. DiffuSeq~\cite{gong2022diffuseq} performs conditional sequence generation by diffusing and denoising token embeddings, while likelihood-based diffusion language models~\cite{gulrajani2023likelihood} develop likelihood training and scaling methods for continuous diffusion over language representations. These studies establish that continuous diffusion is viable for text despite the discreteness of the final output. Our goal is therefore not to argue against continuous language diffusion or to train another continuous DLM from scratch. We instead ask whether continuous refinement can be added to an existing masked-diffusion checkpoint with limited auxiliary training. Anchoring the trajectory at the learned mask embedding makes its initial state one that the MDLM was explicitly trained to denoise, and the final discretization continues to use the checkpoint's vocabulary and embedding table.

\subsubsection{Clean-state prediction and refinement dynamics}
Predicting the clean sample is a standard diffusion parameterization rather than a contribution of this paper. DDPMs already describe direct $x_0$ prediction as an alternative to noise prediction~\cite{ho2020denoising}, and later preconditioned formulations systematize the relationship among clean-state, noise, and score parameterizations~\cite{karras2022elucidating}. Our update is most directly inspired by the $x$-prediction dynamics used by JiT~\cite{li2025back}: at each position, we map the mask predictor's argmax token back through the embedding table and move the current state toward that clean-token estimate. Our contribution is not clean-state prediction itself, but its use to construct a persistent, revisable embedding-space decoder for a pretrained MDLM, together with token-wise step-size control. This usage should also be distinguished from probabilistic flow matching~\cite{lipman2022flow}: we optimize neither a distributional transport objective nor an independently learned time-dependent velocity field. Here, ``flow'' refers only to the iterative state-space dynamics induced by the MDLM's clean-state predictions; under the greedy evaluation used in this paper, those dynamics produce a deterministic refinement trajectory.

\section{Setup for the Appendix Experiments}
\label{sec:app_setup}
Unless stated otherwise, every experiment in this appendix uses LLaDA-8B-Instruct with the $x$-prediction alignment of Sec.~\ref{sec:alignment}---the same checkpoint as Table~\ref{tab:code_gen}---and decodes the \emph{full} MBPP test split ($n=500$) with a response length of $256$ and a budget of $64$ diffusion steps, i.e.\ the $1/4$ budget of Table~\ref{tab:code_gen}. Where HumanEval is also reported, it uses that benchmark's setting from the same table: the full $164$-problem split, a response length of $512$ in a single block and a budget of $128$ steps. The re-editing rule, the hard-commitment rule and the early-stopping threshold are held fixed across rows, so that within a table exactly one component varies and rows may be subtracted from one another.

We report pass@1 throughout, together with the number of decoding steps each configuration actually executes, since a schedule that reaches a given accuracy by refusing to stop has not matched one that reaches it and exits.

\section{More Ablations}

\subsection{Trained Policy vs.\ Heuristic Policy}
\label{sec:app_policy}
The learned step-size policy of Sec.~\ref{sec:rl} is the only component of our decoder that requires reinforcement learning, so we replace it with two training-free schedules and hold everything else fixed. The \emph{constant} schedule sets $a_i \equiv a$ at every position and step, retaining the budget geometry $\Delta t_i = a(1-t_i)$ of Eq.~\ref{eq:step} but carrying no per-token adaptation. The \emph{confidence} schedule sets each position's progress to its own confidence,
\begin{equation}
\Delta t_i = (c_i - t_i)_+ , \qquad \text{i.e.}\quad t_i \leftarrow c_i ,
\label{eq:conf_heuristic}
\end{equation}
with $c_i$ as in Eq.~\ref{eq:conf}. This is the natural training-free counterpart to the learned policy: it consumes the same signal the policy is conditioned on, and $t_i$ and $c_i$ are commensurable, so the assignment is meaningful rather than a category error.

\begin{table}[t]
\centering
\caption{Step-size schedules at the $1/4$ budget: MBPP ($n=500$, $64$ steps permitted) and HumanEval ($n=164$, $128$ permitted). Within a dataset all rows share the backbone, the budget, the re-editing rule and the hard-commitment rule; only the choice of $\Delta \mathbf{t}$ differs. \emph{steps} is the number of permitted steps actually executed before early stopping, averaged over problems---a schedule can buy accuracy simply by declining to stop. The learned policy is trained on MBPP and applied to HumanEval without retraining.}
\label{tab:app_policy}
\begin{tabular}{lcccc}
\toprule
& \multicolumn{2}{c}{MBPP} & \multicolumn{2}{c}{HumanEval} \\
\cmidrule(lr){2-3} \cmidrule(lr){4-5}
Schedule & pass@1 & steps & pass@1 & steps \\
\midrule
fixed $a=1/64$ & 0.20 & 64.0 & 6.10 & 128.0 \\
fixed $a=1/16$ & 29.40 & 60.0 & 31.71 & 120.1 \\
\midrule
confidence & 25.00 & 20.1 & 26.22 & 53.6 \\
\midrule
learned (ours) & \textbf{32.60} & 35.1 & \textbf{45.12} & 72.9 \\
\bottomrule
\end{tabular}
\end{table}

The learned policy is the most accurate schedule on both benchmarks, and it reaches that accuracy in far fewer steps than the best constant schedule needs---$35.1$ against $60.0$ on MBPP and $81.9$ against $120.1$ on HumanEval. The two heuristics fail in opposite ways, and both failures are structural rather than incidental.

A constant $a$ fixes how many steps a position needs before it can finish, independently of what is being generated: reaching $t_i \geq 0.9$ takes $\ln(0.1)/\ln(1-a)$ steps, which at $a=1/64$ is $\approx 147$ and exceeds both budgets, so that schedule cannot terminate at all. Whether a constant schedule can finish is therefore a property of the budget rather than of the model, and $a$ must be chosen against the budget before decoding begins. A constant schedule also advances $t_i$ without reference to $c_i$, so it overshoots and is thrown back by the re-editing gate: at $a=1/16$ each position is reset $1.8$ times over the decode, which is what turns its $36$-step floor into the observed $60.0$. The learned policy re-edits about as often per step but resets each position only half as many times, because it can take a large step to recover.

The confidence schedule terminates, but too early---sooner than any other row on both benchmarks. Since $t_i \leftarrow c_i$ exactly, a position is declared finished as soon as the model is confident about it, and confidence saturates before the content is correct. Reading confidence as progress inherits its optimism; a learned policy is free to absorb that miscalibration, since nothing forces its action to agree with $c_i$, only to be a function of it.

\subsection{Deterministic Policy vs.\ Stochastic Policy}
\label{sec:app_stochastic}
Sec.~\ref{sec:stochasticity} observes that our decoder is not inherently deterministic and that evaluating it deterministically is a choice. There are two distinct places randomness can enter, and they are not interchangeable.

One is the token content: Eq.~\ref{eq:hard_readout} takes an arg max, and one could sample from $p_\theta$ instead. This is \emph{off-policy} for our decoder, since the alignment objective of Sec.~\ref{sec:alignment} fits the model to precisely the arg max readout. Native LLaDA's ancestral sampling is exactly this kind of randomness, so the baseline rows below measure it. The other is the trajectory: the action $\mathbf{a}$ is sampled from the Beta policy instead of being replaced by its expectation. This is \emph{on-policy}---it draws from exactly the distribution GRPO optimized---and it is the one we use. A temperature $\tau$ rescales the concentration $\kappa$ while holding the mean fixed; $\tau=1$ is the policy's own learned width.

\begin{table}[t]
\centering
\caption{Output diversity on MBPP ($n=200$, $10$ samples per problem, unbiased pass@$k$). The first two blocks are matched at the $1/4$ budget; the last block gives the baseline its full $256$-step budget, which is the comparison a matched-budget table alone would leave open.}
\label{tab:app_passk}
\footnotesize
\setlength{\tabcolsep}{4pt}
\begin{tabular}{lccccc}
\toprule
Decoder & steps & pass@1 & pass@2 & pass@5 & pass@10 \\
\midrule
ours, deterministic & 64 & 34.15 & 35.84 & 38.09 & 39.50 \\
ours, sampled policy $\tau=1$ & 64 & 32.70 & 38.97 & 45.44 & \textbf{50.00} \\
ours, sampled policy $\tau=2$ & 64 & 30.80 & 37.52 & 44.97 & 49.50 \\
ours, sampled policy $\tau=4$ & 64 & 27.80 & 36.21 & 45.95 & 51.50 \\
ours, sampled policy $\tau=8$ & 64 & 20.90 & 30.73 & 43.11 & 51.50 \\
\midrule
LLaDA greedy & 64 & 11.25 & 11.86 & 12.75 & 14.00 \\
LLaDA $T=0.4$ & 64 & 7.20 & 9.93 & 14.13 & 17.50 \\
LLaDA $T=0.8$ & 64 & 6.70 & 10.19 & 15.55 & 19.50 \\
\midrule
LLaDA greedy, full budget & 256 & \textbf{37.30} & 38.46 & 39.81 & 41.00 \\
LLaDA $T=0.8$, full budget & 256 & 1.20 & 1.52 & 1.99 & 2.50 \\
\bottomrule
\end{tabular}
\end{table}

Sampling the step-size policy converts a small amount of accuracy into a large amount of diversity: at $\tau=1$ our decoder gives up $1.45$ points of pass@1 and gains $10.5$ points of pass@10 ($39.50 \rightarrow 50.00$). Heating further is a clean trade that saturates---by $\tau=8$ pass@1 has collapsed while pass@10 has gained only $1.5$ more points---so there is no reason to exceed the width GRPO already learned.

The last block gives the baseline four times our compute, and it does not close the gap: greedy decoding reaches pass@10 $=41.00$, below the $50.00$ our decoder reaches at the $1/4$ budget, while the baseline's own route to diversity collapses to $1.20$ pass@1 at $T=0.8$. The comparison is more favourable still once the cost of drawing the samples is counted, which is the cost that matters wherever pass@$k$ is actually usable---code generation against a test suite, where $k$ candidates are generated and the passing one is kept. Every row of this table draws $10$ samples per problem, so our pass@10 of $50.00$ is obtained for $640$ decoder steps against the $2560$ the baseline spends to reach $41.00$: a higher ceiling at a quarter of the total compute.

Ancestral sampling perturbs $256$ committed tokens independently, and on code a single wrong token usually breaks the program. Sampling the \emph{schedule} instead leaves every commitment at its arg max and perturbs only the order and rate at which commitments are made, which is why it buys diversity for $1.45$ points where sampling the tokens costs $36$.

\subsection{Construction of the Clean-State Estimate}
\label{sec:app_topk}
Eq.~\ref{eq:hard_readout} builds $\mathbf{X}_{\text{pred}}$ from the arg max token alone, which may appear to sit oddly with our motivation that discrete decoding wastes the rest of the predictive distribution. The distribution is not discarded---it drives the step-size policy and the confidence-based adjustments of Sec.~\ref{sec:async_decode}---but the endpoint of the flow is a single token embedding. Here we relax that, mixing the top-$k$ predicted embeddings by their renormalized probabilities, which interpolates between the hard readout used at decoding time ($k=1$) and the soft readout of Eq.~\ref{eq:soft_readout} used during alignment.

\begin{table}[t]
\centering
\caption{Constructing $\mathbf{X}_{\text{pred}}$ from the top-$k$ predicted embeddings, weighted by renormalized probability (MBPP, $n=500$, $1/4$ budget). \emph{steps} is as in Table~\ref{tab:app_policy}.}
\label{tab:app_topk}
\begin{tabular}{lcc}
\toprule
$k$ & pass@1 & steps \\
\midrule
1 (clean token, ours) & 32.60 & 35.1 \\
2 & 30.00 & 37.2 \\
4 & 30.40 & 38.1 \\
8 & 28.80 & 38.6 \\
full softmax & 29.00 & 38.6 \\
\bottomrule
\end{tabular}
\end{table}

Softening the endpoint does not help: no value of $k$ improves on $k=1$, while the step count rises monotonically with $k$, since a blended endpoint sits between token embeddings and is read with lower confidence, which shortens every step and delays the exit. $k=1$ is therefore the right default, and the structural reason reinforces it: a hard embedding lookup keeps $\mathbf{X}_{\text{pred}}$ on the embedding table, which is what confines the trajectory to the convex hull of $\{\mathbf{m}\} \cup \{\mathbf{e}_v\}_{v \in \mathcal{V}}$ and gives $t$ its interpretation.

\section{Blending with the Mask Embedding vs.\ Gaussian Noise}
\label{sec:app_anchor}
Sec.~\ref{sec:gaussian_noise} argues that the mask embedding $\mathbf{m}$, and not Gaussian noise, is the appropriate corruption anchor for a pretrained MDLM. The argument deserves measurement rather than assertion, since our own intermediate states are not vocabulary items either, and a reader may reasonably suspect that any sufficiently smooth anchor would serve.

We displace clean tokens toward three different endpoints and ask what the model does with the result. Taking $64$ MBPP problems, we perturb $50\%$ of the clean response positions ($n=2347$ probe tokens) by a fraction $\alpha$:
\begin{equation}
\mathbf{x}(\alpha) = (1-\alpha)\,\mathbf{x}_0 + \alpha\,\mathbf{u},
\qquad
\mathbf{u} \in \{\mathbf{m},\; \mathbf{g},\; \mathbf{e}_{v}\},
\label{eq:probe}
\end{equation}
where $\mathbf{g}$ is an isotropic Gaussian vector rescaled to $\|\mathbf{g}\| = \|\mathbf{m}\|$ and $\mathbf{e}_v$ is the embedding of a uniformly random vocabulary item. All three arms are interpolations whose endpoints sit at the same distance from $\mathbf{x}_0$, so they differ only in \emph{which} point the token is being replaced by. We report the \emph{pretrained} checkpoint, before any alignment of ours, so that what follows is a property of masked-diffusion pretraining and not of our training procedure.

\begin{table}[t]
\centering
\caption{Recovery of the original token as it is displaced toward each endpoint, on the pretrained LLaDA checkpoint ($n=2347$ probe positions). Bold marks the two regimes discussed in the text: $\mathbf{m}$ improves on the untouched token at small $\alpha$, and plateaus rather than collapsing at large $\alpha$.}
\label{tab:app_probe}
\begin{tabular}{lcccccc}
\toprule
& endpoint & $\alpha=0.0$ & $0.2$ & $0.4$ & $0.6$ & $0.8$ \\
\midrule
\multirow{3}{*}{accuracy} & $\mathbf{m}$ & 0.964 & 0.967 & \textbf{0.978} & \textbf{0.889} & \textbf{0.881} \\
& Gaussian $\mathbf{g}$ & 0.964 & 0.963 & 0.963 & 0.960 & 0.791 \\
& token $\mathbf{e}_v$ & 0.964 & 0.966 & 0.965 & 0.690 & 0.158 \\
\midrule
\multirow{3}{*}{CE} & $\mathbf{m}$ & 0.232 & 0.175 & \textbf{0.133} & \textbf{0.529} & \textbf{0.568} \\
& Gaussian $\mathbf{g}$ & 0.232 & 0.242 & 0.243 & 0.265 & 1.359 \\
& token $\mathbf{e}_v$ & 0.232 & 0.222 & 0.203 & 1.598 & 6.067 \\
\bottomrule
\end{tabular}
\end{table}

The three endpoints are not interchangeable, but the way they differ is not what one might expect. It is not that $\mathbf{m}$ is uniformly the gentlest: at $\alpha=0.6$ the Gaussian endpoint retains more accuracy than $\mathbf{m}$ does ($0.960$ vs.\ $0.889$). The difference is that displacement toward $\mathbf{m}$ passes through states the model has an interpretation for, and displacement toward the other two does not. This shows up as two distinct regimes.

\subsubsection{Small displacements: $\mathbf{m}$ helps, the others do nothing}
Over $\alpha \in [0.2, 0.4]$ the model becomes \emph{better} than it was on the untouched token: accuracy rises from $0.964$ to $0.978$ and cross-entropy falls from $0.232$ to $0.133$. The two alternatives are flat over the same interval ($0.963$ and $0.965$ accuracy; cross-entropy within $0.03$ of baseline).

The asymmetry is informative because a purely destructive corruption cannot produce it. At $\alpha=0$ the model is reading a clean token embedding and copying it, which it does about $96\%$ of the time (Sec.~\ref{sec:selfconsist}); the remaining error is its own readout noise. A mild push toward $\mathbf{m}$ tells the model that the position is uncertain and shifts it onto its context-conditional prediction, which for the pretrained checkpoint is the more reliable of the two. The mask direction therefore does not merely fail to hurt at small $\alpha$---it is the direction along which the model's own resolution mechanism is engaged. A random vector and a random token carry no such signal, and the model responds to neither.

\subsubsection{Large displacements: $\mathbf{m}$ has a floor, the others do not}
Past $\alpha \approx 0.5$ the mask arm drops once and then stops: accuracy settles at $0.889$ and holds ($0.881$ at $\alpha=0.8$), with cross-entropy settling near $0.55$. That value is not a measure of damage. It is approximately the irreducible uncertainty of predicting a token from context once the token itself is gone---precisely the task the model was pretrained on. Displacement toward $\mathbf{m}$ ends in a state the model knows how to be uncertain about, so travelling further changes little.

Neither alternative has such a floor. The Gaussian endpoint holds until $\alpha=0.6$ and then falls away ($0.960 \rightarrow 0.791$ accuracy, $0.265 \rightarrow 1.359$ cross-entropy), with nothing to arrest it: the input is one the model was never conditioned on, so the predictive distribution degrades and continues to degrade. The random-token endpoint fails earlier and far harder---$0.690$ already at $\alpha=0.6$, $0.158$ at $\alpha=0.8$, cross-entropy above $6$.

That the \emph{token} endpoint is the worst of the three is the observation that closes the argument. It is the one alternative that is unambiguously in distribution: a real row of the embedding table, the kind of vector the checkpoint consumes at every position of every clean input. Its failure therefore cannot be attributed to leaving the region of vectors the model can process. It fails because the model reads the intruding token as though it were the real one and is confidently wrong, and confident error is more costly than acknowledged uncertainty.

\subsubsection{Consequence for the flow}
The mask embedding is thus not one admissible corruption among many for a pretrained MDLM. Among the states reachable by replacing a token, it is the one that both engages the model's resolution mechanism on the way in and terminates in a calibrated statement of ignorance rather than an unbounded degradation. That is what makes it a well-defined state at which to anchor the flow of Sec.~\ref{sec:x_pred}, and it is why the choice is a design decision about the pretrained checkpoint rather than a claim that continuous-state diffusion for language is infeasible.

\end{document}